\documentclass[11pt,a4paper]{article}
\usepackage[blocks]{authblk}
\usepackage[hyperref]{latell}
\usepackage{times}
\usepackage{latexsym}

\usepackage{microtype}
\usepackage{subcaption}
\usepackage{placeins}
\usepackage[normalem]{ulem}
\usepackage{amsmath,amssymb,amsthm,mathtools}
\usepackage{enumitem}
\usepackage{physics}

\usepackage{booktabs}
\usepackage{multirow}
\usepackage[table]{xcolor}

\aclfinalcopy 

\title{Dependency Parsing Across the Resource Spectrum:\\
Evaluating Architectures on High and Low-Resource Languages}

\author{
Kevin Guan \quad Happy Buzaaba \quad Christiane Fellbaum \\
Princeton University \\
\texttt{\{kevinguan, happy.buzaaba, fellbaum\}@princeton.edu}
}

\date{}

\begin{document}
\aclfinalcopy

\maketitle


\renewcommand{\thefootnote}{\arabic{footnote}}
\setcounter{footnote}{0}

\pagestyle{plain}
\thispagestyle{plain}

\begin{abstract}
Transformer-based models achieve state-of-the-art dependency parsing for
high-resource languages, yet their advantage over simpler architectures in
low-resource settings remains poorly understood. We evaluate four
parsers---the Biaffine LSTM, Stack-Pointer Network, AfroXLMR-large, and
RemBERT---across twelve typologically diverse languages, with a focus on
low-resource African languages. We find that the Biaffine LSTM consistently outperforms
transformer models in low-resource regimes, with transformers recovering their
advantage as training data increases. The crossover falls within a resource
range typical of treebanks for under-resourced languages. Morphological complexity (measured
via MATTR) emerges as a significant secondary predictor of transformers'
relative disadvantage after controlling for corpus size. These results indicate that the Biaffine LSTM may be better suited for syntactic tool development in low-resource regimes until sufficient annotated data is available to leverage the representational capacity of pre-trained transformers.
\end{abstract}

\section{Introduction}

Dependency parsing---the task of identifying grammatical relations between words in a
sentence---is a foundational task in NLP with broad downstream applications,
including machine translation, information extraction, and semantic role
labeling. Transformer-based models such as BERT and its multilingual variants
now dominate dependency parsing benchmarks for high-resource languages, where
abundant pre-training corpora and large treebanks allow these
parameter-rich architectures to perform extremely well
\cite{kondratyuk-straka-2019-75, li2019self, grunewald-friedrich-kuhn-2021-occam}.

This advantage, however, does not transfer uniformly to low-resource languages.
Many languages---particularly African languages such as Xhosa and Wolof, spoken
by millions but severely underrepresented in large text corpora---have dependency
treebanks containing only a few hundred to a few thousand sentences, compared to
tens of thousands available for English
\cite{kumar-jha-sahula-2020-survey, zhou-dakota-kubler-2024-xibe}. Pre-trained
transformers require substantial in-domain data to fine-tune effectively, and the
computational demands of fine-tuning present additional barriers in low-resource
settings \cite{wang-etal-2020-extending, wu-dredze-2020-all}.

We hypothesize that traditional architectures---particularly LSTM-based
parsers---remain competitive with or outperform transformers when training
data is scarce, for three reasons. First, \textbf{generalization}: transformer
models contain orders of magnitude more learnable parameters than LSTM-based
parsers. With limited training data, smaller models may generalize more reliably
and overfit less \cite{gessler-zeldes-2022-microbert,
lindenmaier-papay-pado-2025-efficient}. Second, \textbf{inductive bias}: LSTM
networks encode stronger built-in assumptions about sequential and hierarchical
structure, which may serve as useful guidance when data is scarce
\cite{tran-etal-2018-importance,
dehghani-gouws-vinyals-uszkoreit-kaiser-2018-universal}. Third,
\textbf{stability}: LSTM-based models tend to exhibit more consistent behavior
across training runs, a meaningful advantage when working with small, noisy
datasets \cite{mccoy-frank-linzen-2020-syntax,
dodge-ilharco-schwartz-farhadi-hajishirzi-smith-2020-finetuning}.

To test this hypothesis, we systematically evaluate two LSTM-based parsers
(Biaffine and Stack-Pointer) and two transformer-based parsers (AfroXLMR-large
and RemBERT) across a diverse set of languages spanning a wide
range of treebank sizes. Our central questions are: (1) Do LSTM-based parsers
outperform transformer-based parsers on low-resource languages, while
transformers regain their advantage at higher resource levels? (2) Does the
transformers' relative error disadvantage decrease as training
data grows? We measure performance using labeled and unlabeled attachment
scores (LAS/UAS), and we operationalize the transformer's relative advantage as
its change in error rate normalized against the Biaffine LSTM baseline---a
measure we call relative error rate ($RER$). This measure accounts for the fact
that LAS gains are not comparable across languages and settings with very different
baseline difficulties.

\section{Related Work}
\label{section:related_work}

Most African languages remain severely under-resourced in NLP due to their
near-absence from web-crawled corpora such as Wikipedia and CommonCrawl,
producing a digital language divide in which standard NLP pipelines fail to
generalize to African linguistic contexts
\cite{adelani-2025-nlp-african, bella-2023-digital-language-divide,
nekoto-etal-2020-participatory}.

Recent efforts to close this gap have proceeded along two complementary tracks:
dataset construction and model adaptation. In the data domain, various initiatives have produced word embeddings \cite{adelani-africanlp-resources, bojanowski-etal-2017-enriching}, human-curated parallel corpora, and labeled benchmarks for machine translation \cite{nsumba-etal-2026-salt31, kalejaiye-etal-2025-ibom}, sentiment analysis \cite{muhammad-etal-2023-afrisenti}, and news classification \cite{adelani-etal-2023-masakhanews} across a wide range of languages. On the architectural front, AfriBERTa \cite{ogueji-2022-afriberta} and AfroXLMR \cite{alabi-etal-2022-adapting} adapt multilingual transformer encoders to African morphology and orthography. AfriBERTa is trained from scratch, while AfroXLMR leverages continued pre-training on African-language corpora; both outperform general baselines such as mBERT and XLM-R on African NLP tasks. Masakhane, a grassroots research community, has been central to both efforts, providing named-entity and part-of-speech benchmarks across 20 languages through MasakhaNER \cite{adelani-etal-2022-masakhaner} and MasakhaPOS \cite{dione-etal-2023-masakhapos}, among other contributions.
\section{Data}
\label{sec:data}
Syntactic resources for African languages have developed more slowly than the broader benchmarks described above. Treebanks such as
Wolof-WTB \cite{dione-2019-developing} and Naija-NSC
\cite{caron-2019-surface} establish gold-standard dependency annotations for
individual languages, but broad cross-lingual coverage has remained limited.
AfriSUD \cite{buzaaba2026afrisud} represents a significant step forward,
providing human-annotated treebanks in the
Surface-Syntactic Universal Dependencies (SUD) framework across nine African
languages. This collection forms the basis of the present study, with seven of the nine languages---Swahili, Kinyarwanda, Nigerian Pidgin, Wolof, Yoruba, Xhosa, and Hausa---included in our evaluation.
\paragraph{Languages and resource groupings.}
While the AfriSUD collection provides data for nine languages, we exclude Efik and Runyankore from our
evaluation because pre-trained static word embeddings were unavailable. To span a broader resource spectrum and establish
high-resource baselines, we supplement the remaining seven AfriSUD languages
with five additional SUD treebanks: French, Dutch, Romanian, Turkish, and Afrikaans. This
yields a final evaluation set of twelve languages: Swahili (swa), Kinyarwanda
(kin), Nigerian Pidgin (pcm), Wolof (wol), Yoruba (yor), Hausa (hau), Xhosa
(xho), French (fra), Dutch (nld), Romanian (ron), Turkish (tur), and Afrikaans (afr).
\paragraph{Treebank sources and splits.}
Our treebanks are SUD-converted versions of standard UD releases: French GSD
\cite{abeille-etal-2019-un}, Alpino for Dutch \cite{bouma2001alpino}, Romanian RRT
\cite{barbu-mititelu-irimia-2016-linguistic}, Turkish IMST \cite{sulubacak-etal-2016-universal}, AfriBooms for Afrikaans
\cite{augustinus-etal-2016-afribooms}, and the AfriSUD collection
\cite{buzaaba2026afrisud} for remaining languages. Splits follow an
80--10--10 ratio. Because the French and Dutch training sets are larger than needed to
characterize high-resource behavior, we randomly subsample them to 10{,}000 sentences; dev
and test sets remain unchanged. Sentence counts after subsampling are reported in
Table~\ref{table:treebank_splits}.

\begin{table}[hbt]
  \centering
  \small
  \begin{tabular}{llrrr}
    \hline
    \textbf{Language} & \textbf{Code} & \textbf{Train} & \textbf{Dev} &
    \textbf{Test} \\\hline
    French          & fra & 10{,}000$^{*}$ & 1{,}634 & 1{,}634  \\
    Dutch & nld & 10{,}000$^{*}$ & 1{,}360& 1{,}361 \\
    Romanian        & ron &  7{,}619 &   953 &   952 \\
    Turkish & tur & 4{,}508 &563&564 \\
    Afrikaans       & afr &  1{,}547 &   194 &   193 \\
    Swahili         & swa &  1{,}187 &   149 &   148 \\
    Kinyarwanda     & kin &  1{,}098 &   137 &   137 \\
    Nigerian Pidgin & pcm &  1{,}085 &   135 &   136 \\
    Wolof           & wol &    736   &    92 &    92 \\
    Yoruba          & yor &    694   &    87 &    87 \\
    Hausa           & hau &    690   &    86 &    86 \\
    Xhosa           & xho &    369   &    46 &    46 \\\hline
  \end{tabular}
  \caption{Treebank splits (sentence counts). $^{*}$French and Dutch training sets
    subsampled from 13{,}074 and 10{,}882, respectively, to 10{,}000; dev/test sets unchanged.}
  \label{table:treebank_splits}
\end{table}

\paragraph{Morphological complexity.}
To examine whether typological properties of the target language modulate
architecture-level differences in parsing performance, we include morphological
complexity as a covariate in our statistical models. We operationalize complexity
via the Moving-Average Type--Token Ratio (MATTR) \cite{covington2010cutting},
which corrects the well-known sensitivity of raw type--token ratio to text length
by averaging type--token ratios over a sliding window. Languages with richer morphology produce more distinct surface forms per window,
yielding higher MATTR values. We use the gold tokenization provided by our treebanks and a window size of 500 tokens. MATTR scores are computed over each language's
training split and reported in Table~\ref{table:mattr}.

The resulting rankings align with established characterizations of morphological complexity. Xhosa, Turkish, and
Kinyarwanda---languages with productive agglutinative morphology
\cite{probert2019comparison, briscoe-carroll-1993-generalised, nzeyimana-2024-low}---rank highest,
reflecting the large inventory of distinct surface forms generated by
concatenative affixation. Romanian ranks fourth, consistent with its retention
of nominal case morphology distinguishing it from its Romance relatives
\cite{maiden2021oxford}. At the other extreme, Nigerian Pidgin and Afrikaans score
lowest, in line with the morphological reduction characteristic of pidginization and Afrikaans's well-documented morphological simplification
from Dutch \cite{mcwhorter2001simplest, donaldson1993afrikaans}.

\begin{table}[hbt]
\centering
\small
\begin{tabular}{lccc}
\hline
\textbf{Language} & \textbf{Tokens} & \textbf{MATTR} & \textbf{$z$-score} \\\hline
Xhosa           &   6{,}282 & 0.800 & $+$2.241 \\
Turkish         &  46{,}239 & 0.732 & $+$1.389 \\
Kinyarwanda     &  26{,}700 & 0.673 & $+$0.658 \\
Romanian        & 173{,}946 & 0.634 & $+$0.175 \\
Swahili         &  29{,}781 & 0.626 & $+$0.076 \\
Dutch           & 154{,}187 & 0.609 & $-$0.130 \\
Yoruba          &  10{,}422 & 0.586 & $-$0.409 \\
French          & 244{,}944 & 0.570 & $-$0.617 \\
Wolof           &  15{,}415 & 0.567 & $-$0.651 \\
Hausa           &  13{,}874 & 0.566 & $-$0.662 \\
Nigerian Pidgin &  26{,}489 & 0.541 & $-$0.974 \\
Afrikaans       &  39{,}199 & 0.531 & $-$1.095 \\\hline
\end{tabular}
\caption{MATTR scores and standardized values, computed over
training splits with window size $w = 500$.}
\label{table:mattr}
\end{table}

\section{Models}
We evaluate three primary dependency parsing architectures: the Biaffine LSTM parser \citep{dozat-manning-2017-deep}, transformer-based biaffine parsers, and the Stack-Pointer Network \citep{ma-etal-2018-stack}. The biaffine parsers are implemented using the SuPar Python package \citep{zhang-parser}, and the Stack-Pointer parsers are implemented using the NeuroNLP2 repository \citep{ma-neuronlp2}.

\paragraph{Biaffine LSTM.}
The deep biaffine attention parser \citep{dozat-manning-2017-deep} is our graph-based baseline and the de facto standard for neural dependency parsing. It encodes sentences with a BiLSTM and scores arcs via a biaffine function, decoding the optimal tree with the Chu-Liu/Edmonds algorithm \citep{chu-1965-shortest, edmonds-1967-optimum}. We use this model with static FastText embeddings.

\paragraph{Stack-Pointer Network.}
The Stack-Pointer Network \citep{ma-etal-2018-stack} is a transition-based parser with a BiLSTM-CNN encoder that decodes top-down and depth-first via a pointer network, incorporating ancestor and sibling features for higher-order context. This approach avoids the directional bias of traditional transition parsers while remaining more efficient than MST-based decoding.

\paragraph{AfroXLMR-large.}
AfroXLMR-large \citep{alabi-etal-2022-adapting} is a domain-adapted variant of XLM-R \citep{conneau-etal-2020-unsupervised} that continues pre-training on 17 African languages, yielding richer representations for the morphologically complex languages in our evaluation set. It is paired with SuPar's biaffine parsing head and fine-tuned end-to-end.

\paragraph{RemBERT.}
RemBERT \citep{chung2020rethinking} is a multilingual transformer encoder that decouples input and output embeddings, allocating a larger input layer to better handle morphologically rich languages. Like AfroXLMR-large, it is paired with a biaffine head in SuPar and fine-tuned jointly during training.

\section{Experiments}

\paragraph{Hyperparameter optimization.}
We tune each model via grid search with a fixed seed. To keep the search
tractable, high-resource languages are downsampled to 5{,}000 training
sentences, and models are trained at 25\% of the full epoch budget with halved
early-stopping patience. For the Biaffine LSTM we search over learning rate,
decay rate, and decay steps (12 runs per language); for the transformer-based
models (AfroXLMR-large and RemBERT) we search over learning rate, warmup steps,
and the ratio between the encoder and task-head learning rates (18 runs); and
for Stack-Pointer we additionally tune unknown-word replacement probability and
gradient clipping (24 runs). All other hyperparameters are kept at default values as defined by the respective implementation libraries.

\paragraph{Word embeddings.}
The Biaffine LSTM and Stack-Pointer models use 300-dimensional pre-trained
static word vectors. For languages in the official FastText distribution
\citep{grave2018learning} we use vectors trained on CommonCrawl and Wikipedia.
For African languages not covered there, we fall back to embeddings provided by Adelani's AfricaNLP
resources repository \citep{adelani-africanlp-resources}. As noted in
Section~\ref{sec:data}, Efik and Runyankore are excluded from our analysis
because no pre-trained vectors could be located for either language. Static
vectors are concatenated with character-level representations (CNN or LSTM) and
gold POS tag embeddings. The transformer-based models do not use static vectors; they
generate contextualized subword representations internally.

\paragraph{Training protocol.}
Final models are trained on the full data with the best hyperparameters. Biaffine models run for up to 1{,}000 epochs (patience 100);
Stack-Pointer models run for up to 600 epochs (patience 60). In practice, no
model approached its epoch ceiling before early stopping triggered. Transformer
models on high-resource languages (Romanian, Dutch, French) are additionally
capped at 100 epochs to balance computational costs, as these models require fewer passes to converge due to their large volumes of data. Each configuration is trained five times with different
random seeds, and we report mean UAS and LAS across runs to reduce sensitivity
to initialization.

\section{Results}

\subsection{Parsing Performance}

Tables~\ref{table:las_results} and~\ref{table:uas_results} report mean LAS and
UAS with standard deviations across five training runs for all four
architectures and twelve languages. Overall, performance scales with training set
size, though the relative ordering of architectures varies by language. The
transformer models outperform the Biaffine LSTM on higher-resource
languages, while the LSTM is competitive with or superior to both transformers
on several lower-resource ones. The Stack-Pointer parser is the weakest
architecture across nearly all languages and metrics.

\begin{table*}[hbt]
\centering
\small

\begin{subtable}{\textwidth}
\centering
\begin{tabular}{lcccc}
\toprule
\textbf{Language} & \textbf{Biaffine LSTM} & \textbf{AfroXLMR-large} &
\textbf{RemBERT} & \textbf{Stack-Pointer} \\\midrule
French (fra)           & $93.51 \pm 0.13$ & $95.81 \pm 0.05$ & $95.53 \pm 0.08$ & $92.41 \pm 0.05$ \\
Dutch (nld)             & $91.07 \pm 0.12$ & $94.55 \pm 0.15$ & $94.50 \pm 0.13$ & $90.39 \pm 0.21$ \\
Romanian (ron)        & $87.23 \pm 0.18$ & $90.61 \pm 0.12$ & $90.51 \pm 0.06$ & $87.41 \pm 0.12$ \\
Turkish (tur)           & $67.35 \pm 0.52$ & $72.76 \pm 0.56$ & $73.07 \pm 0.49$ & $63.88 \pm 0.28$ \\
Afrikaans (afr)       & $85.85 \pm 0.31$ & $88.90 \pm 0.28$ & $87.75 \pm 0.19$ & $85.91 \pm 0.61$ \\
Swahili (swa)         & $79.48 \pm 0.43$ & $78.48 \pm 0.49$ & $75.77 \pm 0.53$ & $73.10 \pm 0.95$ \\
Kinyarwanda (kin)     & $76.16 \pm 0.30$ & $76.21 \pm 0.43$ & $75.16 \pm 0.94$ & $74.76 \pm 0.35$ \\
Nigerian Pidgin (pcm) & $70.56 \pm 0.51$ & $72.12 \pm 0.42$ & $71.81 \pm 0.52$ & $68.65 \pm 0.77$ \\
Wolof (wol)           & $79.63 \pm 0.28$ & $78.06 \pm 0.76$ & $76.99 \pm 0.42$ & $76.27 \pm 0.76$ \\
Yoruba (yor)          & $77.71 \pm 0.66$ & $77.75 \pm 0.81$ & $76.13 \pm 0.42$ & $75.33 \pm 0.45$ \\
Hausa (hau)           & $79.66 \pm 0.61$ & $80.75 \pm 0.63$ & $79.75 \pm 1.25$ & $78.55 \pm 0.60$ \\
Xhosa (xho)           & $70.38 \pm 0.67$ & $66.00 \pm 0.80$ & $61.08 \pm 1.72$ & $58.20 \pm 1.30$ \\
\bottomrule
\end{tabular}
\caption{Mean Labeled Attachment Score (LAS) $\pm$ standard deviation.}
\label{table:las_results}
\end{subtable}
\vspace{1.5em} 
\begin{subtable}{\textwidth}
\centering
\begin{tabular}{lcccc}
\toprule
\textbf{Language} & \textbf{Biaffine LSTM} & \textbf{AfroXLMR-large} &
\textbf{RemBERT} & \textbf{Stack-Pointer} \\\midrule
French (fra)           & $95.53 \pm 0.08$ & $97.30 \pm 0.03$ & $97.17 \pm 0.07$ & $94.69 \pm 0.08$ \\
Dutch (nld)             & $93.85 \pm 0.16$ & $96.34 \pm 0.12$ & $96.37 \pm 0.07$ & $93.24 \pm 0.22$ \\
Romanian (ron)        & $92.45 \pm 0.14$ & $95.38 \pm 0.08$ & $95.26 \pm 0.04$ & $92.01 \pm 0.11$ \\
Turkish (tur)           & $75.98 \pm 0.23$ & $82.30 \pm 0.37$ & $82.86 \pm 0.30$ & $72.40 \pm 0.37$ \\
Afrikaans (afr)       & $88.74 \pm 0.31$ & $91.37 \pm 0.33$ & $90.42 \pm 0.01$ & $88.68 \pm 0.59$ \\
Swahili (swa)         & $88.57 \pm 0.37$ & $89.39 \pm 0.22$ & $87.54 \pm 0.57$ & $84.92 \pm 0.82$ \\
Kinyarwanda (kin)     & $86.44 \pm 0.24$ & $86.59 \pm 0.26$ & $86.11 \pm 0.54$ & $85.71 \pm 0.32$ \\
Nigerian Pidgin (pcm) & $80.41 \pm 0.72$ & $82.13 \pm 0.30$ & $81.83 \pm 0.44$ & $78.12 \pm 0.48$ \\
Wolof (wol)           & $85.79 \pm 0.16$ & $84.88 \pm 0.76$ & $84.19 \pm 0.47$ & $82.83 \pm 0.77$ \\
Yoruba (yor)          & $85.46 \pm 0.41$ & $85.97 \pm 0.67$ & $84.25 \pm 0.39$ & $83.69 \pm 0.35$ \\
Hausa (hau)           & $92.30 \pm 0.49$ & $92.32 \pm 0.29$ & $91.81 \pm 0.74$ & $91.47 \pm 0.74$ \\
Xhosa (xho)           & $81.70 \pm 0.52$ & $76.78 \pm 1.07$ & $72.78 \pm 1.69$ & $69.80 \pm 1.44$ \\
\bottomrule
\end{tabular}
\caption{Mean Unlabeled Attachment Score (UAS) $\pm$ standard deviation.}
\label{table:uas_results}
\end{subtable}
\caption{Parsing performance results across five seeds for LAS and UAS.}
\label{table:combined_results}
\end{table*}

We quantify architecture-level differences using the \textit{relative error
rate} ($RER$). Let $\text{LAS}_{\text{Biaffine LSTM}}(\ell)$ and
$\text{LAS}_{\text{TF}}(\ell)$ denote the labeled attachment scores of the
Biaffine LSTM baseline and a transformer comparison model on language $\ell$.
We define:
\begin{equation}
  RER_{\text{LAS}}(\ell)
    = \frac{\text{LAS}_{\text{Biaffine LSTM}}(\ell) - \text{LAS}_{\text{TF}}(\ell)}
           {100 - \text{LAS}_{\text{Biaffine LSTM}}(\ell)}
  \label{eq:rer}
\end{equation}
and likewise for unlabeled attachment  ($RER_{\text{UAS}}$). We prefer $RER$
over raw score differences for two reasons. First, absolute LAS or UAS gains
are confounded with baseline difficulty: a 2-point gain at a baseline of 95
is not comparable to the same gain at a baseline of 70, as the remaining
headroom differs substantially. Second, because attachment scores are bounded
at 100, raw differences are heteroscedastic across the training-size axis, with
low-resource languages exhibiting lower baselines and higher variance;
normalizing against the remaining error stabilizes variance across languages.
A value $RER_{\text{LAS}} > 0$ indicates that the comparison model incurs
more errors than the Biaffine LSTM, while $RER_{\text{LAS}} < 0$ indicates fewer
errors.

\begin{figure}[hbt]
    \centering
    \begin{subfigure}{0.48\textwidth}
        \centering
        \includegraphics[width=\linewidth]{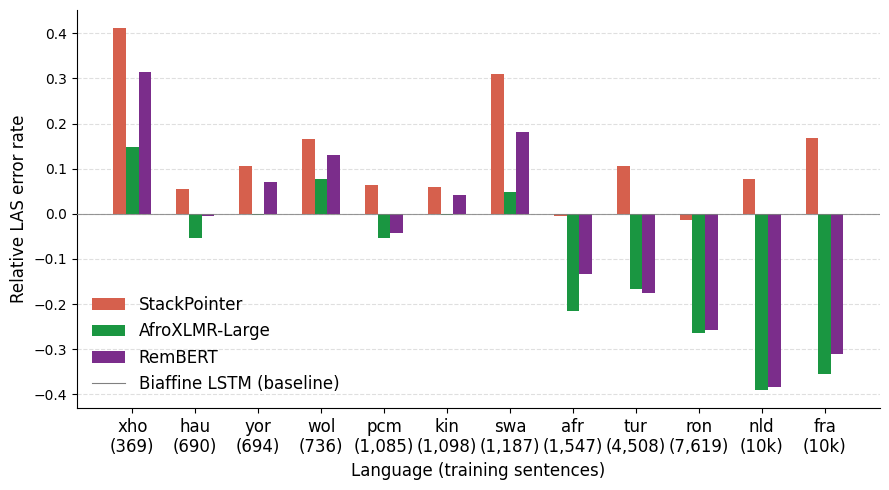}
        \caption{$RER_{\text{LAS}}$}
        \label{fig:err_las}
    \end{subfigure}
    \hfill
    \begin{subfigure}{0.48\textwidth}
        \centering
        \includegraphics[width=\linewidth]{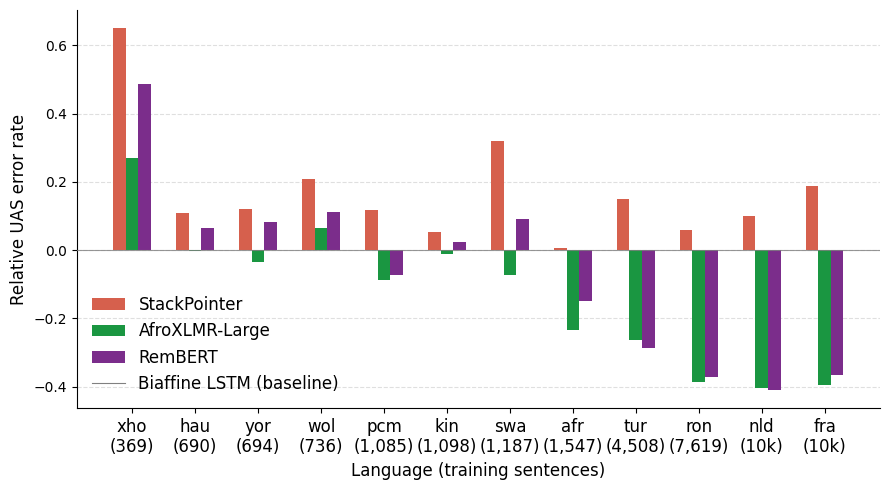}
        \caption{$RER_{\text{UAS}}$}
        \label{fig:uas_err}
    \end{subfigure}
    
    \caption{$RER$ vs. sentence count. $RER< 0$ indicates fewer errors than Biaffine LSTM baseline.}
    \label{fig:combined_err}
\end{figure}

Tables~\ref{table:relative_error_las} and~\ref{table:relative_error_uas} report
$RER_{\text{LAS}}$ and $RER_{\text{UAS}}$ for each comparison model.
Both transformer models show $RER_{\text{LAS}}, RER_{\text{UAS}} < 0$ (fewer errors than the LSTM) on French,
Romanian, and Afrikaans, but $RER > 0$ on Xhosa, Wolof, and---for
RemBERT---Swahili. Looking only at $RER_{\text{LAS}}$, the pattern of LSTM's competitiveness extends further. AfroXLMR shows positive $RER_{\text{LAS}}$ on Swahili ($+0.049$) in addition to Xhosa ($+0.148$) and Wolof ($+0.077$). RemBERT's label-level underperformance is even more extensive: beyond Swahili ($+0.181$), Wolof ($+0.130$), and Xhosa ($+0.314$), it records positive $RER_{\text{LAS}}$ on Yoruba ($+0.071$) and Kinyarwanda ($+0.042$). The Stack-Pointer parser consistently underperforms, with $RER > 0$ on nearly every language. Figure~\ref{fig:combined_err} plots $RER_{\text{LAS}}$ and
$RER_{\text{UAS}}$ for each non-baseline model across languages ordered by
training set size.
\begin{table}[hbt]
\centering
\small
\begin{tabular}{lccc}
\toprule
\textbf{Language} & \textbf{AfroXLMR} & \textbf{RemBERT} & \textbf{Stk-Ptr} \\\midrule
French          & $-0.354$ & $-0.311$ & $0.169$  \\
Dutch           & $-0.390$ & $-0.384$ & $0.076$  \\
Romanian        & $-0.265$ & $-0.257$ & $-0.014$ \\
Turkish         & $-0.166$ & $-0.175$ & $0.106$  \\
Afrikaans       & $-0.216$ & $-0.134$ & $-0.004$ \\
Swahili         & $0.049$  & $0.181$  & $0.311$  \\
Kinyarwanda     & $-0.002$ & $0.042$  & $0.059$  \\
Nigerian Pidgin & $-0.053$ & $-0.042$ & $0.065$  \\
Wolof           & $0.077$  & $0.130$  & $0.165$  \\
Yoruba          & $-0.002$ & $0.071$  & $0.107$  \\
Hausa           & $-0.054$ & $-0.004$ & $0.055$  \\
Xhosa           & $0.148$  & $0.314$  & $0.411$  \\
\bottomrule
\end{tabular}
\caption{$RER_{\text{LAS}}$ compared to the Biaffine LSTM
baseline. Values $RER_{\text{LAS}} < 0$ favor the
comparison model; values $RER_{\text{LAS}} > 0$ favor the LSTM.}
\label{table:relative_error_las}
\end{table}
\begin{table}[hbt]
\centering
\small
\begin{tabular}{lccc}
\toprule
\textbf{Language} & \textbf{AfroXLMR} & \textbf{RemBERT} & \textbf{Stk-Ptr} \\\midrule
French          & $-0.396$ & $-0.367$ & $0.188$  \\
Dutch           & $-0.405$ & $-0.410$ & $0.100$  \\
Romanian        & $-0.388$ & $-0.372$ & $0.059$  \\
Turkish         & $-0.263$ & $-0.286$ & $0.149$  \\
Afrikaans       & $-0.234$ & $-0.149$ & $0.006$  \\
Swahili         & $-0.072$ & $0.090$  & $0.320$  \\
Kinyarwanda     & $-0.011$ & $0.024$  & $0.054$  \\
Nigerian Pidgin & $-0.088$ & $-0.072$ & $0.117$  \\
Wolof           & $0.064$  & $0.113$  & $0.208$  \\
Yoruba          & $-0.035$ & $0.083$  & $0.122$  \\
Hausa           & $-0.003$ & $0.064$  & $0.108$  \\
Xhosa           & $0.269$  & $0.487$  & $0.650$  \\
\bottomrule
\end{tabular}
\caption{$RER_{\text{UAS}}$ compared to the Biaffine LSTM
baseline. Values $RER_{\text{UAS}} < 0$ favor the comparison model; values
$RER_{\text{UAS}} > 0$ favor the LSTM.}
\label{table:relative_error_uas}
\end{table}

\subsection{Training Data Scaling}

To test whether $RER$ is systematically related to training set size, we fit
linear mixed models predicting $RER_{\text{LAS}}$ and $RER_{\text{UAS}}$ as a
function of log training sentences, with language as a random intercept. We
restrict this analysis to the two transformer models, since the Stack-Pointer
parser is consistently outperformed by the LSTM baseline, showing $RER > 0$ on nearly every language.

\begin{figure}[t]
    \centering
    \begin{subfigure}{0.48\textwidth}
        \centering
        \includegraphics[width=\linewidth]{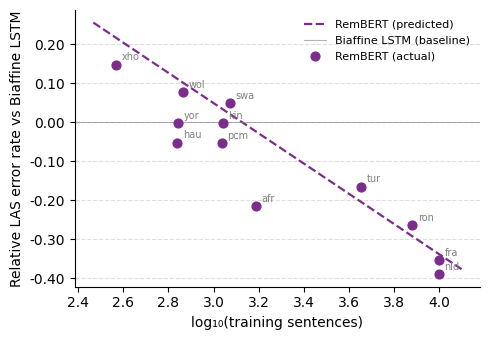}
        \caption{$RER_{\text{LAS}}$: RemBERT vs. Biaffine LSTM}
        \label{fig:mlm_las_rembert}
    \end{subfigure}
    \hfill
    \begin{subfigure}{0.48\textwidth}
        \centering
        \includegraphics[width=\linewidth]{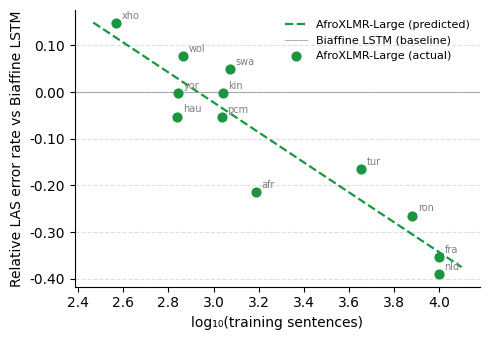}
        \caption{$RER_{\text{LAS}}$: AfroXLMR-large vs. Biaffine LSTM}
        \label{fig:mlm_las_afro}
    \end{subfigure}

    \caption{$RER_{\text{LAS}}$ as a function of training size. Plots for $RER_{\text{UAS}}$ are omitted for brevity, although the trends are qualitatively identical, with $RER_{\text{UAS}}$ decreasing as training data grows.}
    \label{fig:mlm_full_comparison}
\end{figure}

Both models show a statistically significant negative relationship
between training size and $RER$---$RER$ decreases as more
data becomes available. For LAS, the slope on \texttt{log\_train} is
$\hat{\beta} = -0.388$ ($p < .001$) for RemBERT and $\hat{\beta} = -0.321$
($p < .001$) for AfroXLMR-large; the corresponding UAS slopes are steeper at
$\hat{\beta} = -0.496$ and $\hat{\beta} = -0.398$ (both $p < .001$). Spearman
correlations between training size and mean $RER$ are negative and significant across all model--metric combinations, corroborating this pattern under minimal distributional assumptions. For RemBERT, $\rho_{\text{LAS}} = -0.830$ ($p < .001$) and $\rho_{\text{UAS}}
= -0.869$ ($p < .001$); for AfroXLMR-large, $\rho_{\text{LAS}} = -0.844$
($p < .001$) and $\rho_{\text{UAS}} = -0.935$ ($p < .001$). Fitted curves for LAS are shown in
Figure~\ref{fig:mlm_full_comparison}. 

The fitted curves cross $RER = 0$---the point at which a transformer's
predicted error equals the LSTM's---at the sentence counts summarized in
Table~\ref{table:crossovers}. AfroXLMR-large reaches the crossover
at around 830-860 sentences, while RemBERT requires approximately
1,310--1,340. Several languages in our evaluation fall close to these
thresholds: Hausa (690), Yoruba (694), and Wolof (736) sit just below the
AfroXLMR-large crossover, while Swahili (1,187), Nigerian Pidgin (1,085), and Kinyarwanda (1,098)
fall between the estimates.

\begin{table}[hbt]
\centering
\small
\begin{tabular}{llcc}
\hline
\textbf{Model} & \textbf{Metric} & \textbf{$\log_{10}$(train)} & \textbf{Sents.} \\\hline
AfroXLMR-large & LAS & 2.931 & 854   \\
RemBERT        & LAS & 3.127 & 1{,}339 \\
AfroXLMR-large & UAS & 2.924 & 839   \\
RemBERT        & UAS & 3.118 & 1{,}311 \\\hline
\end{tabular}
\caption{Estimated training-set sizes at which fitted $RER$
curve crosses $RER = 0$, i.e.\ where transformers' predicted
error equals the Biaffine LSTM's.}
\label{table:crossovers}
\end{table}

\subsection{Morphological Complexity}

To assess whether morphological complexity explains residual variance in
architecture-level differences beyond what training size captures, we
extend the mixed models by adding MATTR as a predictor alongside log training
sentences, and compare model fit with and without this term.

\begin{figure}[t] 
    \centering
    \begin{subfigure}[b]{0.48\textwidth}
        \centering
        \includegraphics[width=\linewidth]{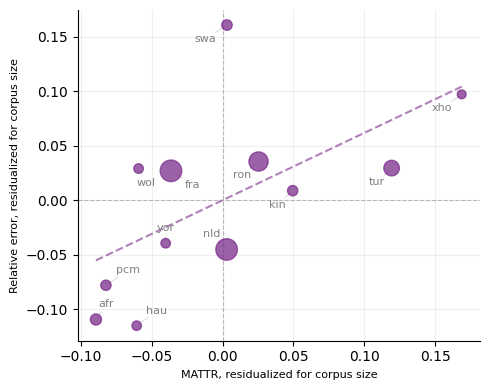}
        \caption{RemBERT: $RER_{\text{LAS}}$}
        \label{fig:partial_rembert_las}
    \end{subfigure}
    \hfill
    \begin{subfigure}[b]{0.48\textwidth}
        \centering
        \includegraphics[width=\linewidth]{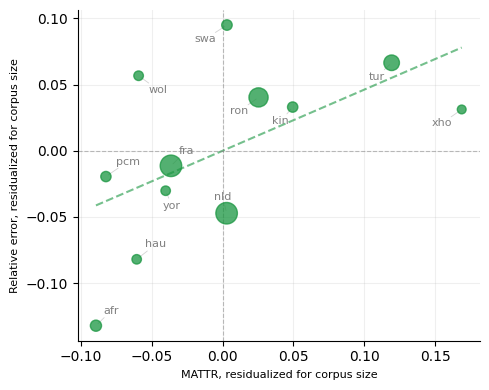}
        \caption{AfroXLMR-large: $RER_{\text{LAS}}$}
        \label{fig:partial_afro_las}
    \end{subfigure}
    \caption{Partial regression of $RER_{\text{LAS}}$ on MATTR for RemBERT (top) and AfroXLMR-large (bottom), after residualizing both variables for training size. Partial regression plots for $RER_{\text{UAS}}$ are omitted for brevity, although the relationship of $RER_{\text{UAS}}$ with morphological complexity is consistent with that of $RER_{\text{LAS}}$ and leads to the same conclusions.}
    \label{fig:partial_regressions_combined}
\end{figure}

For RemBERT, adding MATTR significantly improves fit on both LAS
($\chi^2(1) = 5.44$, $p = .020$) and UAS ($\chi^2(1) = 8.65$, $p = .003$),
with positive \texttt{mattr\_z} coefficients of $\hat{\beta} = 0.050$
($p = .023$) and $\hat{\beta} = 0.065$ ($p = .002$) respectively: higher
morphological complexity is associated with a positive shift in
$RER$, even after controlling for data size. Similarly, for AfroXLMR-large, both the LAS model ($\chi^2(1) =
4.51$, $p = .034$; $\hat{\beta} = 0.037$, $p = .043$) and the UAS model
 ($\chi^2(1) = 9.89$, $p = .002$; $\hat{\beta} =
0.048$, $p = .001$) show a significant effect. The magnitude of the MATTR effect is somewhat smaller for AfroXLMR than for RemBERT on both metrics. It is possible that AfroXLMR's African-language pretraining partially offsets the elevated $RER$ that complex morphology imposes on transformer models, though this was not formally tested. In all extended
models, the negative effect of training size on $RER$ remains
significant. Figure~\ref{fig:partial_regressions_combined}
shows partial regression plots in which variables have been residualized for
training size, isolating the complexity effect.

\section{Discussion}

\subsection{Scaling Behavior Across Architectures}

Across both transformer models, $RER$ decreases as training corpus size grows, with transformers achieving $RER \leq 0$ once treebanks
reach roughly 800--1,400 training sentences. This pattern is consistent with
standard fine-tuning dynamics: multilingual pre-trained models carry useful
cross-lingual syntactic representations, but adapting those representations to
a specific treebank's annotation conventions requires sufficient supervision.
With little training data, fine-tuning is underconstrained---the model lacks sufficient signal to effectively map its pre-trained representations to the specific structural conventions of the target treebank. The Biaffine LSTM, learning its representations entirely from
the treebank, avoids this problem; its inductive biases toward locality and
sequential composition suit the short-range dependencies that dominate
treebank data. As supervision increases, transformers gain the signal
needed to resolve this ambiguity, $RER$ crosses zero, and their cross-lingual
transfer becomes advantageous.

The estimated crossover points are practically significant: a substantial
fraction of African UD treebanks fall within the 839--1,339 sentence range,
meaning the LSTM remains the stronger choice for many under-resourced languages
in practice. $RER_{\text{UAS}}$ shows a steeper scaling trajectory than
$RER_{\text{LAS}}$. We attribute this to structural attachment being more
typologically general and therefore more transferable via pre-training, while
label assignment depends more heavily on language-specific annotation
conventions that neither scale nor pre-training fully resolves.

\subsection{Morphological Complexity and the Low-Resource Penalty}

Even after controlling for corpus size, morphological complexity predicts a
consistent positive shift in $RER$ for both transformer models. Morphologically
rich languages produce a larger inventory of distinct inflected forms, compounding the
difficulty of learning an accurate parser from limited data. The effect is
weaker for AfroXLMR-large---its MATTR coefficient is roughly 1.35 times
smaller than RemBERT's---which is consistent with its continued pre-training on
African-language text providing richer coverage of inflectional
paradigms, partially offsetting the morphological component of $RER$.

In absolute terms the MATTR effect is modest, and interpretation is complicated
by Xhosa's status as the smallest treebank and most extreme
MATTR outlier, giving it disproportionate leverage over the complexity
estimates. We also note that MATTR captures lexical diversity rather than
morphological complexity directly. More principled operationalizations---such
as morpheme-per-word ratios or paradigm entropy---would provide a stronger test in future work. While lemmatizing the treebanks could theoretically mitigate this morphological penalty, robust lemmatizers for many of these languages remain unavailable.
\subsection{Stack-Pointer Underperformance}
The Stack-Pointer parser shows $RER > 0$ on nearly every language, suggesting
that its architectural complexity is counterproductive in this data regime. Its
top-down, depth-first decoding introduces error propagation, since
mistakes at high nodes cascade to descendants---an effect that is
exacerbated when data is scarce. The pointer network decoder also
adds substantially more parameters than the Biaffine LSTM, increasing the risk
of overfitting on small treebanks, and the higher-order sibling and grandparent
features that distinguish this architecture depend on prior parsing decisions
that may themselves be unreliable under low-resource conditions.

\subsection{Broader Limitations}

Several factors limit the generalizability of our conclusions. (1) Our treebanks
differ in domain and genre, and some of the observed $RER$
differences may reflect genre mismatch rather than resource level. (2) For
non-transformer models, the quality of word embeddings introduces a
confound that remains unquantified in our crossover estimates, since
higher-resource languages benefit from embeddings trained on larger corpora.
(3) Our transformer baselines are all multilingual---languages with monolingual
pre-trained encoders available may show different crossover behavior. (4) Typological
differences in word order and non-projectivity affect parsing difficulty
independently of resource level and cannot be fully separated from it in our
sample. (5) The test set for our lowest-resource language contains fewer than
50 sentences, limiting the statistical precision of reported means. A larger and
more diverse language sample would provide more robust grounding
for our claims.

\section{Conclusion}

This work examined whether transformer-based parsers maintain their advantage
over recurrent architectures when training data is scarce. Across twelve typologically diverse languages spanning a wide range of resource levels, our results consistently show $RER > 0$ for both AfroXLMR-large and RemBERT on the
lowest-resource languages, while $RER$ decreases with training data until transformers achieve $RER \leq 0$ at higher resource levels.
Statistical analysis confirms a significant negative relationship between training set size and $RER$, and morphological complexity---operationalized via MATTR---emerges as a secondary predictor of elevated $RER$: higher morphological richness compounds the transformer's error disadvantage, an effect that is attenuated for AfroXLMR-large, likely due to its pre-training on African-language text. These findings suggest that data scarcity and
morphological richness jointly constrain the effectiveness of multilingual transformer fine-tuning in low-resource conditions.

The LAS scores on the lowest-resource languages remain lower than most downstream applications require, but they nonetheless carry
practical value. Imperfect parsers may still assist with annotation,
reducing the clerical burden on human annotators and accelerating
treebank development. Parsing accuracy also functions as a sensitive diagnostic
of the overall NLP infrastructure available for a language. Taken together,
these results suggest that the Biaffine LSTM may be better suited for syntactic tool development in under-resourced languages---for which $RER > 0$ favors it over transformer-based models---until sufficient annotated data is available to leverage the representational capacity of large pre-trained models.

\paragraph{Future work.}
Several directions would strengthen and extend these findings. The precise $RER = 0$ crossover could be localized by systematically
subsampling larger treebanks to produce controlled, within-language training
sets of varying sizes. Whether the crossover generalizes beyond dependency
parsing to structurally simpler tasks such as POS tagging and NER, and whether
zero-shot or few-shot cross-lingual transfer can push it toward lower resource
levels, are open questions. The evaluation should also be extended to morphologically rich agglutinative languages---Turkic languages, Japanese, Korean, and Indigenous languages of the Americas---which are not represented in the current study, and to architectures
such as the TreeCRF parser \citep{zhang-etal-2020-efficient}, which has been
reported to perform well under low-resource conditions. Finally, more principled
measures of morphological complexity---such as information-theoretic metrics---would allow a cleaner test of
whether the elevated $RER$ for transformers is specifically driven by
inflectional disambiguation load, and evaluating these architectures on lemmatized data, where available, could test if neutralizing inflectional variance mitigates this MATTR confound.

\section*{Acknowledgments}
This work was supported by the Princeton Language and Intelligence (PLI) Seed Grant Program. We thank Janet Pierrehumbert for the conversation that led to the central hypothesis on which this work is based.

\bibliographystyle{acl_natbib}
\bibliography{latell}

\end{document}